\documentclass[journal]{IEEEtran}
\usepackage{cite}
\usepackage[utf8]{inputenc}
\usepackage{times}
\usepackage{epsfig}
\usepackage{graphicx}
\usepackage{amsmath}
\usepackage{amssymb}
\usepackage{cite}
\usepackage{soul}
\usepackage{float}
\usepackage{algorithm}
\usepackage[noend]{algpseudocode}
\usepackage{color}
\usepackage{hyperref}

\begin{document}

\title{3D Scanning: A Comprehensive Survey}

\author{Morteza~Daneshmand,
        Ahmed~Helmi,
        Egils~Avots,
        Fatemeh~Noroozi,
        Fatih Alisinanoglu,
        Hasan~Sait~Arslan,
        Jelena~Gorbova,
        Rain~Eric~Haamer,
        Cagri~Ozcinar,
        Gholamreza~Anbarjafari
\thanks{M. Daneshmand, A. Helmi, E. Avots, F. Noroozi, H. S. Arslan, J. Gorbova, R. E. Haamer, and G. Anbarjafari are with iCV Research Lab, Institute of Technology, University of Tartu, Tartu, Estonia. e-mail: shb@icv.tuit.ut.ee}
\thanks{F. Alisinanoglu and G. Anbarjafari are also with Department of Electrical and Electronic Engineering, Hasan Kalyoncu University, Gaziantep, Turkey.}
\thanks{C. Ozcinar is with Trinity College Dublin (TCD), Dublin 2, Ireland. e-mail: ozcinarc@scss.tcd.ie}
}

\maketitle

\begin{abstract}
This paper provides an overview of 3D scanning methodologies and technologies proposed in the existing scientific and industrial literature. Throughout the paper, various types of the related techniques are reviewed, which consist, mainly, of close-range, aerial, structure-from-motion and terrestrial photogrammetry, and mobile, terrestrial and airborne laser scanning, as well as time-of-flight, structured-light and phase-comparison methods, along with comparative and combinational studies, the latter being intended to help make a clearer distinction on the relevance and reliability of the possible choices. Moreover, outlier detection and surface fitting procedures are discussed concisely, which are necessary post-processing stages.
\end{abstract}


\IEEEpeerreviewmaketitle

\section{Introduction}
Automated scene interpretation is a topic of major interest in photogrammetry, Remote Sensing ({RS}), and computer vision. 3D shape measurement, also known as range imaging and depth sensing, plays an important role in many applications. However, the mainstream methods do not provide measurements accurate and dense enough for a reliable 3D reconstruction \cite{zhang2013unambiguous,moeslund2001survey,tong2012scanning,anbarjafari20173d}.\\
3D scanning of scenes and multiple objects is time-consuming, and thus bottlenecked by manual labor. Nevertheless, since the associated processing techniques are under constant improvements, the latter drawback has been substantially alleviated in the last few years. It should be noted that usually, when the object of interest has a distinctive surface, photogrammetric techniques are favored.\\
On the other hand, there are not enough free benchmark databases available for testing the associated algorithms and procedures. It should be noted that photogrammetry may not be as efficient as Total Station ({TS}) and ground-based Interferometric Synthetic-Aperture Radar ({InSAR}) systems, due to accuracy and response-time requirements.\\
In~\cite{hattab2017differential}, a differential 3D scanning procedure was proposed, in each iteration of which, the designer creates a 3D model, prints it using a 3D printer, edits it, and then scans it back to apply the changes. The true changes made to the model are automatically detected based on the reference model, which are chosen based on the coefficients related to the printing accuracy, aiming at minimizing the accumulation of errors. The method does not work well on shiny and translucent objects, and as 3D printers are slow, this type of iterative workflow is limited by the number of iterations.\\
In~\cite{zhang2015turning}, methods of 3D on-road object extraction from Mobile Laser Scanning ({MLS}) data were represented, e.g. for road surface, pavement cracks, road markings and manhole covers, where the samples were collected at driving speeds.\\
In~\cite{el2015comparison}, a comparison of analytical aerial photogrammetry, i.e. B/W stereo photographs, Time-Of-Flight ({TOF}) {LS}-based measurement mechanic, {TS} based on ground control points and rover with {GPS} surveys was provided in terms of generating Digital Terrain Models ({DTM}). Reportedly, {TS} led to the best accuracy, although demonstrated a very slow performance in data collection, with {LS} being second, which demands complex data analyses. Aerial photogrammetry required expensive equipment to be viable, and Real-Time Kinematic ({RTK}) {GPS} was the most unreliable one.\\
In~\cite{fan2016automated}, an easy-to-use motorized and automated system was developed that scans 3D scenes in a reasonable time frame, which can be utilized for viewing scalable multi-object scenes, as well as for path planning purposes. The resolution they achieved was 0.1~mm on a variety of different objects.\\
Due to the fact that 3D scanning with Light Detection And Ranging ({LiDAR}) is expensive, in~\cite{bi2017using}, it was attempted to achieve comparable results on topography mapping of a fault zone with a low-cost digital camera mounted on a {UAV}. They used the Agisoft PhotoScan software~\cite{verhoeven2011taking} for processing the images in order to reconstruct the ground. Comparisons with {LiDAR} showed nearly no difference in accuracy, but the Structure-From-Motion ({SFM})-based photogrammetry method yielded a denser point cloud.\\
In~\cite{remondino2011uav}, an evaluation of Unmanned Aerial Vehicle ({UAV})- and photogrammetry-based 3D scanning methods was provided. It revealed that {UAV}s have a lower operational cost, where the task can usually be performed by two operators, but often carry small or medium format cameras, which is why they require more images to reach the same accuracy as conventional methods.\\
In~\cite{sun2016rba}, an Reduced Bundle Adjustment ({RBA}) method was proposed for reducing the computational complexity and memory cost of the task of Bundle Adjustment ({BA}). It uses both oblique and nadir camera data to fixate their orientations and positions. The accuracy of the proposed method is slightly poor, but the efficiency is significantly improved.\\
One of the important applications of {RS} is in exploring, recording or documenting objects, landscapes and heritage sites~\cite{remondino2011heritage}. New sensors and multi-resolution 3D scanning and visualization methodologies developed in the foregoing context are aimed at purposes such as mapping and conservation.\\
One of the applications of {PaRS} is in improving the speed and accuracy of progress reporting, where a construction site is scanned regularly, and the scans are compared to each other, as well as to the planned outcome, in order to measure the amount of work done quantitatively, and help make management and control decisions~\cite{el2008integrating}.\\
A homemade 3D scanner consisting of a camera and a projector was built in~\cite{gorte2013camera}, aiming at relatively high levels of accuracy and precision, as well as a compact and dexterous setup. As a case-study, the system was utilized for monitoring a flume under construction.\\
As demonstrated in~\cite{schmidt2003full}, 3D image correlation photogrammetry, possibly in combination with stroboscopic illumination~\cite{schmidt1985stroboscopic} or high-speed video cameras~\cite{tsukada2001direct,oguma2011high}, has applications in full-field dynamic displacement, deformation, stress and strain measurement, e.g. for ionic polymeric and dielectric elastomer artificial muscles~\cite{shahinpoor1998ionic,anderson2012multi}, and presents a higher robustness and dynamic range than other full-field imaging approaches such as 3D Electronic Speckle Pattern Interferometry ({ESPI})~\cite{lokberg1987electronic} or moir\'{e} topography~\cite{takasaki1970moire}.\\ In~\cite{schmidt2003full}, pulsed laser was used to investigate the behavior of a flywheel in a spin pit at a maximum speed of 35,000 Revolutions per Minute ({RpM}), and a notched rubber dogbone sample was pulled to failure in 38~ms, where numerous strain maps were obtained.\\
While displaying a point cloud obtained through photogrammetry, it should be decided which points are visible from a given viewpoint, where surface- and voxel-based methods~\cite{bernardini1999ball,kuzu2004volumetric} have been proposed for Hidden Point Removal ({HPR}), which have been investigated in~\cite{alsadik2014visibility}, along with issues such as guidance with synthetic images~\cite{bernardini1999ball}, camera network design~\cite{alsadik2013automated,granshaw2010close} and gap detection and classification~\cite{impoco2004closing}.\\
Photogrammetry can make use of the geometric data present in data resources shared through social media such as Facebook and Instagram, in order to calibrate and recover information about important facts, events and locations, such as burial places, with possible changes occurring from the time of taking an image to another one potentially introducing further challenges~\cite{stylianou2014images}.\\
In~\cite{geibig2016compact}, a compact 3D imaging radar based on Frequency-Modulated Continuous-Wave ({FMCW})-driven frequency-scanning antennas was proposed for a small and lightweight radar-system, which is equipped for catching two orthogonal 2D pictures. Angular and parallel determination of these pictures are adequate to supply vital data about the field in basic circumstances. Moreover, a fundamental algorithmic approach was proposed for joining the two pictures to a pseudo 3D model of the landscape. As the compass time per check is under 1~ms, and every 3D picture is developed from only two filters, the frame-rates achievable with this framework design are appropriate for real-time utilization.\\
In~\cite{zolotukhin20133d}, an algorithm has been proposed that makes it conceivable to fabricate a 3D model of a miniaturized object on the premise of a stereo combination of pictures acquired by a {REM}. The parallax in the stereo match is required for revising the performance of the algorithm. The basic grid is computed through Random Sample Consensus ({RANSAC}). The algorithm is adapted for work with vast magnification coefficients, and can utilize both parallel and point-of-view projection models.\\
In~\cite{huang2013generative}, a hybrid generative statistical framework was proposed for automatic extraction and reconstruction of building roofs from {ALS} point clouds.\\
In~\cite{8kite}, a method was proposed which allows to obtain {DEM}s and the corresponding orthophotography of surfaces larger than 3 km$^{2}$ at {VHR}. They used two delta kites of 4 and 10 km$^{2}$ and a SONY NEX-5N camera~\cite{udin2014assessment}, which met a compromise between weight, image quality and geometry, and cost. In the proposed method, the kit flight angle stability was properly estimated by taking several flights with different conditions and flight-line lengths. One of the limitations of the proposed method is that the route of flight should be open and clear.\\
The importance of icing thickness measurements of transmission lines for preventing possible resulting accidents has been investigated in~\cite{huang2015transmission} through photogrammetry. The proposed system consists of a High-Resolution ({HR}) camera, a laser range finder and an inertial measurement unit, which has led to a flexible and accurate performance along with a conventional fixed terminal.\\
In~\cite{chen2017photogrammetry}, photogrammetry was used for 3D reconstruction of oil paintings. They obtained several sets of orthophotomaps to create 3D models of oil paintings through UV inkjet printing~\cite{calvert2001inkjet}. They used the collinearity equation, i.e. the condition equation, in order to represent the image details, followed by reconstructive accuracy analysis. The contours and hypsometric tints of the model were determined by using spot color swatches in four colors for a UV printer, in order to print 1-10-layer spot color swatches. Then {RGB} image segmentation and section extraction were performed. The results showed that digital photogrammetry can effectively establish a colorful 3D model for oil paintings.\\
Coastal monitoring is fundamental to studying dune and beach behavior related to natural and anthropogenic factors, as well as coastal management programs. Various tools have been applied for the foregoing task, including {LiDAR}, satellite images and {TLS}.\\
This paper reviews various 3D scanning technologies, including close-range, aerial, {SFM} and terrestrial photogrammetry, along with mobile, terrestrial and airborne {LS}, followed by {TOF}, structured-light and phase-comparison approaches, and some of the existing comparative and combinational studies. Last but not least, outlier detection and surface fitting processes will be briefly discussed as well.\\
The purpose of this paper is to help grasp an understanding of the state-of-the-art in the area of 3D scanning. Therefore, the articles to be reviewed have been picked up from the list of publications by the most highly accredited conferences and journals in the last four years up to the time the paper has been written, as well as a few older, but most frequently cited, seminal, ones. 
\section{General Examples of Utilizing Photogrammetry}
A few illustrative examples of usage of photogrammetry will be presented in this section, followed by discussions on more specific types of it in the upcoming ones.
\subsection{Fast Implementation of a Radial Symmetry Measure}
\begin{figure*}[t]
	\centering
	\includegraphics[width=\textwidth]{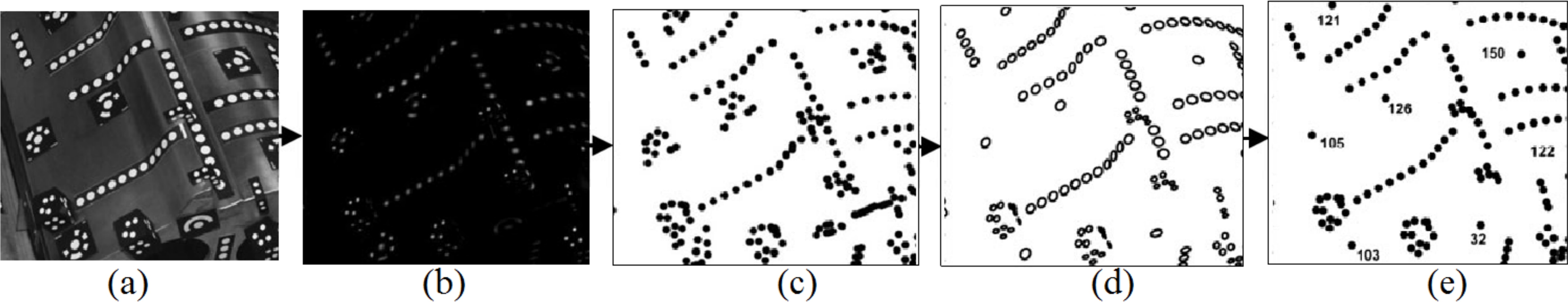}
	\caption{The general pipeline of the radial symmetry measure system proposed in~\cite{fucinos2013fast}: (a)~Input image, (b)~Locations of radial symmetry, (c)~Local maxima of symmetry, (d)~Elliptical fitting and rejection, (e)~Coded target decoding. The figure has been taken from~\cite{fucinos2013fast}.}
	\label{1}
\end{figure*}

In~\cite{fucinos2013fast}, a parallel implementation of two algorithms for target detection and recognition was reported, which led to competitive results, but were computationally expensive. They parallelized both of the algorithms on {CPU} using OpenMP~\cite{openmp}, and implemented the slower parts, e.g. target detection, on {GPU}. One of the observed results was that increasing the size of the image, the execution time on {GPU} increases less than either of the two executions on {CPU}. Fig.~\ref{1} represents the general pipeline of the proposed system.
\subsection{Survey of Natural Areas of Special Protection}
In~\cite{drosos2013use}, a pair of diapositives were scanned using a photogrammetric scanner, and processed with the Leica Photogrammetry Suite ({LPS})~9. They used the scans to evaluate ecological degradation, where Geographic Information System ({GIS}) was taken into account in the survey of natural areas of special protection, with the aesthetic forest Kouri Almyrou Magnisias as a case-study.
\subsection{Relative Pose Measurement of Satellite and Rocket}
Image-based measurements can be used for calculating the relative pose of satellite-rocket separations~\cite{wang2017relative}. In the case the camera parameters are known, the camera has been installed on the rocket launcher, and more than six cooperation signs have been set on the satellite, the latter are captured in real-time, and the pose estimation task is handled by combined virtues of the Efficient Perspective-n-Point ({EPnP})~\cite{yu2009videometrics} and Orthogonal Iteration ({OI})~\cite{lu2000fast} methods, the latter being referred to as LHM as well, which is abbreviated from the last names of the authors, i.e. Lu, Hager and Mjolsness. In the aforementioned study, an accuracy assessment framework has been proposed as well, which has verified the reliability of the foregoing strategy in terms of estimating the relative position and attitude.
\section{Close-Range Photogrammetry ({CRP})}
{CRP} is one of the most prevalent variants of photogrammetry, whose applications will be discussed in this section.
\subsection{Automatic Camera Calibration}
Obtaining {CRP} measurements requires calibration of the digital camera based on sensor modeling, which can be achieved through self-calibration. It has been reviewed based on targeted planar arrays and target-less scenes in~\cite{luhmann2016sensor}.\\
In a {CRP}-image-based 3D measurement context, the camera calibration parameters for {DSLR} cameras~\cite{kim2013digital}, as well as high-end or consumer-grade ones, need to be calculated independently from the characteristics of the scene at hand, which can be performed automatically based on coded targets~\cite{fraser2013automatic}. However, the latter study suggests that a set of considerations are required for an optimally accurate and reliable calibration procedure, which concern either the self-calibration functional model or practical issues, and among others, include the effect of image scales on lens distortions, color imaging and chromatic calibration, focal plane distortion, 3D object space control, image point distribution and interior orientation stability.
\subsection{Line Detection and Matching}
In~\cite{lopez2013detection}, a novel approach for line matching was proposed, which is designed specifically to achieve high performances in industrial environments, characterized by low textured objects and uncontrolled illumination conditions. They extracted the line segments via phase congruency and continuity criteria, and then matched them individually according to their local appearances and photometric properties, without resorting to any prior knowledge about scene or camera position. The total number of extracted lines was reportedly close to the real number of perceived lines, avoiding fragmented lines and overlapping segments. The line matching procedure exhibited a high level of robustness against changes of rotation, scale, perspective and brightness.
\subsection{Forest Analysis}
{CRP} has been widely used for analyzing and monitoring different characteristics pf forests, which will be elaborated in sthis section.
\subsubsection{Plant Diversity and Surface Fuel Structure}
In~\cite{bright2016introducing}, a {CRP} approach was introduced and discussed for measuring the 3D structure of understory vegetation and woody debris, and to test the utility of photogrammetric points for distinguishing and predicting understory fuels and plant diversity. They compared height data derived from {CRP} and field measurements of fuelbed depth, and evaluated the utility of photogrammetry-derived data for separating and classifying 10-cm scale fuel types, plant species, and plant types by studying the area located in Air Force Base ({AFB}) in northwestern Florida.\\
A total of 57 different species were observed, with an average of 19 species per plot and a standard deviation of 6 species across plots.  Digital imagery was processed with the photogrammetry software PhotoModeler Scanner~\cite{photomodeler}. Using the {Smart Match} feature-based method, stereo pairs were created by automatically detecting and matching pixels of similar texture and color between overlapping paired images. The statistical and classification analyses were such that the overall classification accuracy was computed as 100 minus the out-of-bag estimate of error rate. Overall quantity and allocation difference were calculated using the diffeR package in R. The Field-measured heights were often much greater than photogrammetric heights.\\
Overall classification accuracies of fuel type, species, and plant type classifications were 44, 39 and 44 percent, respectively. Patterns and distributions of point-intercept measurements of fuelbed depth and photogrammetric maximum height were similar, but there were discrepancies. The results indicated that {CRP} has a potential for yielding fine-scale measurements of understory fuels and plants. Advantages of {CRP} in the foregoing context include its ability to create a permanent record of understory vegetation and fuels that would support retrospect analyses, and for the calibration of human interpreters.
\subsubsection{Mapping Eroded Areas on Mountain Grassland}
In~\cite{1mapping}, the problem of erosion in mountain areas has been investigated, and a method for mapping and monitoring eroded areas has been proposed for selected scenes, using {CRP} data. In order to automatically extract eroded and grass-covered, i.e. non-eroded, parts of selected 3D point clouds of the scene, they presented an approach applied on object-based image classification with automatic threshold optimization, using the Excess Green Vegetation Index. For classification, the raster-based feature ExG was used, which  characterizes the segments in terms of spectral properties. The proposed method resulted in overall accuracies between 90.7\% and 95.5\%, comparing the automatically extracted eroded areas with manual detection.
\subsection{A Simulation Tool for the {SRT}}
In~\cite{buffa2016simulation}, a {CRP} approach was used to measure the self-load deformations of the {SRT} primary reflector from its optimal shape by using a near real-time {CRP} system with cameras installed in fixed positions. They developed a simulation pipeline to realistically reproduce and evaluate photogrammetric surveys of large structures. The simulation environment consisted of the following items:
\begin{itemize}
	\item A detailed description of the {SRT} model, the measurement points and the camera parameters;
	\item A tool capable of generating realistic images accordingly to the above model;
	\item A self-calibrating {BA} procedure to evaluate the performance in terms of Root-Mean-Square Error ({RMSE}) of the camera configurations.
\end{itemize}
AICON 3D Studio~\cite{aicon3d}, MicMac~\cite{micmac} and MATLAB~\cite{matlab} have been used in the simulation process. The study has solved the self-calibrating {BA} for a pair of exemplifying scenarios, and evaluated the results by specifically comparing AICON 3D Studio and MicMac. A quantitatively similar performances in terms of {RMSE} and estimate of interior camera parameters, having measured the target positions with different accuracies, depending on their radial distance from the reflector vertex.
\subsection{River Surface Water Topography Mapping at Sub-mm Resolution and Precision}
High-precision and high-density measurements of surface water topography in laboratory rivers are useful in hydro-morphodynamic studies for analyzing energy, momentum, and mass balances. In~\cite{han2014river}, wax powder was used as seeds to generate {HR} and high-accuracy {DEM}s of a flowing river water surface. They collected stereo-pair photos of the wax powder, and then post-processed them using 3DM Analyst~\cite{3dmanalyst} and 3DM Calibcam~\cite{3dmcalibcam} commercial software. They achieved instantaneous sub-mm vertical and horizontal resolutions of surface water topography.
\subsection{City Modeling}
In~\cite{singh2014new}, a procedure was described for making large-scale 3D city models using video image sequences. Initially, the buildings were filmed in order to obtain a 360 coverage. Afterward, the video sequences were split into frames, and the camera positions for a selected number of frames were obtained by finding the fundamental matrix. The point clouds of the buildings were created with the Agisoft Photoscan Pro software~\cite{themistocleous2015methodology}, which utilizes the {SFM} algorithm. The 3D model was created by generating a wireframe and adding shading and textures to create a photorealistic representation of a scene. The compositions of the 3D models was created in the Autodesk{\textregistered} 3ds Max{\textregistered} software~\cite{reinhart2009experimental}. The results of the foregoing study showed that 3D model creation from video sequences is simple and cost-effective.
\section{Aerial Photogrammetry}
The potential of {UAV}s for imagery tasks in the context of {PaRS} has attracted attentions from numerous research communities for decades. According to~\cite{colomina2014unmanned}, their competitiveness lies in the fact that they have combined low-cost capturing devices with advances computer-vision and robotic systems, which lead to cm-level accuracies.\\
{RS} and {RPAS} are used in different contexts such as environmental studies, cultural heritage, civil engineering, forestry and cartography. MicMac~\cite{micmac} is a popular open-source software which allows to process the images resulting from {RPAS} for generating georeferenced
information, which can be manipulated or visualized under a {GIS} environment. In~\cite{mic_mac}, an open-source {GIS} application was proposed based on MicMac, and tested on both natural and urban environments, obtaining {HR} databases. Moreover, {UAV}-based photogrammetry can be utilized in contexts such as landslide cut, whose results are suitable for repair and management.\\
The rest of the subjects related to aerial photogrammetry will be discussed from different perspectives in what follows.
\subsection{The {ISPRS} Benchmark}
The {ISPRS} Benchmark on 3D Semantic Labeling~\cite{rottensteiner2014results} was gathered specifically for fostering research on automated analysis. However, only a few approaches have been evaluated on it due to several classes having quite similar geometric behavior, making the classification on the database rather challenging.\\
In~\cite{nex2015isprs}, airborne oblique images were obtained on four cardinal and nadir directions, along with {UAV} images on both nadir and oblique directions, as well as terrestrial images, consisting of both convergent and redundant ones, of two test areas. They prepared databases for two tasks, namely, dense image matching evaluation and Image orientation evaluation.\\
In~\cite{feat_sem}, a framework was presented for semantic labeling on a per-point basis, using collections of spherical and cylindrical neighborhoods, as well as the spatial bins as the basis for a multi-scale geometric representation of the surrounding of each point in the point cloud. In addition, baseline results were provided on the {ISPRS} database.
\subsection{Feature Extraction and Matching}
In~\cite{sun2014l2}, an efficient feature extraction and matching implementation was presented for large images in large-scale aerial photogrammetry experiments, which intended to overcome memory issues with {SIFT}. The red-black tree structure~\cite{herlihy2003software}, the tree key exchange method and segment matching were used to improve the efficiency, having in mind that near-zero pitch/roll angles might make the application unrealistic.
\subsection{Accurate Optical Target Pose Determination}
In~\cite{cucci2016accurate}, a novel optical target design was developed with the goal of determining its center, as well as its distance from {UAV}-based mapping elevations relatively accurately. They determined the target pose and the projection of the circle center, followed by evaluating the results based on real-world data, which demonstrated remarkable accuracies, noting that further improvements could be achieved correcting for the exposure induced bias.
\subsection{Survey of Historical Heritage}
Investigation and observation of historical heritage sites can broadly benefit from the virtue of aerial photogrammetry, which will be discussed in this section.
\subsubsection{Pre-Hispanic Wall Painting}
In~\cite{lucet20133d}, the use of photogrammetry for creating a 3D reconstruction of the {Las Higueras} architectural structure has been described. The photographic survey was made with a Canon EOS 5D Mark II camera with a resolution of 21.1~Mega-pixels. Altogether, 964 pictures were taken, of which two thirds were used. The image processing was done using the MicMac software~\cite{micmac}. The generated 3D model produced acceptable results regarding correct survey of paintings and conservation treatment, with a resolution of 4 pixels per mm.
\subsubsection{Historical Buildings of the City of Strasbourg}
Both small- and large-scale cartography can be achieved through {UAV} photogrammetry, which is helpful for tasks demanding short response times, e.g. observing or monitoring buildings which may be inaccessible to conventional terrestrial devices~\cite{murtiyoso2017acquisition}. In the latter study, based on Sensefly Albris~\cite{senseflyalbris} and the DJI Phantom 3~\cite{djiphantom3} Professional {UAV}s, the related acquisition and processing techniques have been studied as well, which involve calibration, flight planning and data management. The case-studies considered included two historical buildings of the city of Strasbourg, namely, a part of the Rohan Palace façade and the St-Pierre-le-Jeune Catholic church, as well as the Josephine Pavilion. In~\cite{murtiyoso2017acquisition}, it has been concluded that small sensors are associated with less reliable calibration parameters, where dense matching is more challenging, especially in the presence of noise.
\subsection{Street-level Modeling}
This section discusses the problem of street-level modeling on the basis of aerial photogrammetry, as one of its substantial applications.
\subsubsection{Steep Surfaces}
Regularly monitoring the boundaries of a territory including steep surfaces, e.g. mountainsides, quarries, landslides or river basins, is amongst the important applications of aerial photogrammetry, which can be achieved with a decent level of details using lightweight {UAS}s, having recently led to higher-quality results taking advantage of inclined images under oblique photogrammetry frameworks, rather than relying on strictly vertical images~\cite{casella2016modelling}. In the latter study, a large sandpit has been considered as a case-study, with linear, radial and semi-circular strips as different flight configurations, where 300 control points were captured using a topographic {TS}. The evaluation of the acquired point clouds was performed based on their densities, as well as their differences with given checkpoints.
\subsubsection{Building Abstraction}
City model reconstruction is one of applications of aerial photogrammetry and image-based 3D reconstruction. Multi-view stereo has been shown to be a viable tool for the foregoing task. However, it results in dense, but noisy and incomplete, point clouds when applied to aerial images. In~\cite{build_abstract}, a fully-automatic pipeline was applied in order to generate a simplified mesh from a given dense point cloud. Specifically, they employed the results of the prior works on interpreting the estimation of a dense depth map as a labeling problem. The depth map, which is turned into a height map, serves as an intermediate step for the construction of a simplified mesh which is as close as possible to the data. Major geometric structures were maintained while clutter and noise were removed.
\subsubsection{Effects of Image Orientation and Ground Control Point Distributions}
In~\cite{carvajal2016effects}, the procedure of generating a 3D model of a landslide using the {UAV} platform Microdrones md4-20041 was described. The vehicle took photos while flying through predefined flight paths. All the photogrammetric operations were carried out with the Photomodeler Scanner V2012~\cite{brutto2012computer} software. The {DEM}s and the adjusted mosaic of images were used as input data to the orthorectification process, which consisted in reprojecting the mosaic based on the morphology of the {DEM}, and then performing interpolation in order to fill the holes in the {DOM}s.
\subsection{Topographic Monitoring}
According to~\cite{barlow2017kinematic}, {UAV} photogrammetry can generate 3D models of cliff faces, for kinematic stability analysis purposes, with a point density and accuracy that is similar to those produced using {TLS}, but with a reduction in equipment costs.\\
Moreover, in~\cite{uav_coastal}, a relatively low-cost method of using {UAV} was proposed, which includes aerial triangulation with camera calibration and subsequent model generation, being mostly automated. They demonstrated high-accuracy and high-quality results, analyzing two very sensitive test areas on the Portuguese northwest coast.\\
The above study was successfully replicated in~\cite{seasonal_dune} on the Ravenna beach dune system. The methodology can also be applied in multidisciplinary scientific studies such as those in the context of economics and management, in order to evaluate environmental changes from a monetary perspective, or in biology, for the sake of investigating the changes in vegetation and habitats.
\subsection{Determining Fault Planes}
The accuracy and precision of digital photogrammetry and field observation procedures utilized by {UAV}s may be increased by non-metric oblique and vertical camera combination, as well as by properly configuring ground control points, according to~\cite{amrullah2016product}, which have shown to perform equally well in producing {DEM} and {LiDAR}, where the fault plane is found through detecting the cross-section and interpreting the extreme height of terrain changes.\\
In~\cite{zhang2013digital}, a practical framework has been established for resolving early aerial photogrammetric processing in the Tanlu fault zone throughout China East, whose activities have led to some strong earthquakes. Digital photos obtained by high-precision special scanners and the elements of relative orientations of image pairs were obtained by theoretical analysis. Then whole digital photos were processed by using the Inpho photogrammetry software~\cite{inpho}. The horizontal and vertical deformation values were the main quantitative parameters in the morphotectonics research.\\
By processing the above aerial photo images, {DEM}, {DOM} and large-scale topographical map covering, the interesting areas were acquired, based on special maps such as profile map, 3D landscape map and gradient map, for the sake of analyzing fault and morphotectonics.
\subsection{Improving Height Accuracy Using a Multi-class System}
In~\cite{7multicamera} a method for aerial photogrammetry was proposed, which improves the height accuracy. They collected multiple groups of stereo images obtained by a multi-camera system with 60\% overlap between adjacent stations and 90\% overlap between simultaneous images. They compared the results of using a single camera with those of a four-camera system. Their experiments resulted in the conclusion that the height accuracy by one group of stereo images is worse than using redundant stereo images.
\section{Structure-From-Motion ({SFM}) Photogrammetry}
{HR} topographic surveying based on triangulation of the data acquired at known poses of cameras or control points is not only financially considerably costly, but also inconvenient, due to the inaccessibility of some fields of interest, which are the fundamental drawbacks of {TLS}, and consequently, the {GPS}~\cite{westoby2012structure}. Therefore, the latter seminal work proposes the {SFM} algorithm for alleviating the underlying shortcomings of classical photogrammetric systems, which automates the aforementioned procedures through applying highly redundant {BA} processes~\cite{triggs1999bundle} according to the matching feature points observed on overlapping bunches of images having been acquired using a computer-level digital camera, thereby achieving decimeter-level vertical accuracies on land areas with complex topographies and with various scales. The proposed method has been successfully tested on an exposed rocky coastal cliff, a breached moraine-dam complex and a glacially-sculpted bedrock ridge in the original study.\\
From an ecological perspective, one of the applications of {SFM} photogrammetry is in surveying vegetation structures of drylands, and accordingly, their manner of functioning, which has been performed using a small {UAS} on a semi-arid ecosystem in~\cite{cunliffe2016ultra}. Ultra-fine 3D height models with cm-level spatial resolutions were obtained from landscape levels. Above-ground biomass was predicted from canopy volume with rather high coefficients of determination, i.e. $r^{2}$ ranging from 0.64 to 0.95, which proved to be sensitive to vegetation structures, being useful for capturing changes caused by environmental conditions in spatially or temporally discontinuous canopy covers. It is worth noticing that drylands overall constitute more than 40\% of the whole terrestrial surface.
\section{Terrestrial Photogrammetry}
The most essential applications of terrestrial photogrammetry are in topological monitoring, whose examples will be presented in what follows.
\subsection{Coastal Projectors}
In~\cite{12cpro}, a robust coastal projector monitoring system called C-Pro was introduced, which uses terrestrial photogrammetry to project a photograph to a georeferences plane. A mathematical formulation was proposed where the roll and pitch rotation angles were computed using an approximation of the horizon curve. The proposed method allows to make an accurate projection of the coastal photograph on the georeferences plane, even missing internal orientation parameter. The main restriction of the proposed method is that it has been developed only for imaging systems containing horizon curves.
\subsection{Landslides}
Photogrammetry is useful for observing landslides and shape changes of scenes~\cite{roncella2014landslide}. A competitive system of the aforementioned type has been developed in the latter study, which compensates for possible attitude variations automatically. The system consists of two reflex cameras and a local computer which receives the preliminarily captured data, and makes the {DSM} of the scene to be sent to the host computer. The Mont de la Saxe landslide has been considered as a case-study.
\section{General Examples of Usage of {LS}}
{LS} offers highly accurate 3D topographic point clouds that have traditionally not been provided by competing {RS} technologies, which will be generally discussed in this section, followed by the investigation of more specific types of it in the upcoming ones.
\subsection{Heritage Monitoring}
Heritage monitoring is a fundamental use-case of {LS}. The associated examples will be presented in what follows.
\subsubsection{Statues of Michelangelo}
In~\cite{levoy2000digital}, a system was introduced for digitizing the shape and color of large statues, with the statues of Michelangelo as case-studies. They proposed new methods for representing, viewing, aligning, merging, and viewing large 3D models using a laser triangulation scanner. The costs of equipment shipping and the protection considerations have been reported as the main underlying difficulties.
\subsubsection{The Guyue Bridge}
Under severe environmental conditions, in combination with historic evidence, precise 3D models obtained through {LS} can be used to observe, and possibly speculate previous stages of, heritage sites, which has been performed in order to approximate the original dimensions and scales of the Guyue Bridge, which had been constructed on A.D 1213 by the Southern Song Dynasty in Yiwu, China, as well as to capture the folding arch, i.e. the transitional stage from beam to curved arch, in~\cite{lu2015application}.
\subsection{Mining}
Different 3D {LS} technologies utilized for mining were discussed in~\cite{fengyun2013status} based on the scanner applications elaborated on the basis of domestic situation, consisting of 3D reconstruction and measurement of open pit, headframe deformation monitoring, mining subsidence monitoring, construction land reclamation regulatory, difficult arrival region survey of coal gangue dump, landslide monitoring and deformation monitoring of underground mined area. They also discussed five aspects of possible future developments, which should be able to provide some measure of protection for construction of digital mines.\\
In~\cite{guo20163d}, two solutions were described for roadway modeling: The projection method and Poisson equation, experiments of roadway modeling based on which were carried out. The compared results of the experiment showed that the roadway model established by the Poisson equation method was smoother than that of the cylinder projection method, and that the modeling time of Poisson equation method was much larger than that of the cylinder projection method. Both methods could also be further applied to similar tunnel engineering problems, such as subway tunnels.
\subsection{Configuration Modeling for a Constrained Elastica Cable}
In~\cite{du2014configuration}, explanatory solution formulas as elliptic functions were picked up to depict spatial cable setup in a cylindirical shape framework. The system was set up as indicated by Saint-Venant guidelines~\cite{toupin1965saint}. Based on the {OCC} solid modeling kernel~\cite{occ}, utilizing the Gauss-Legendre quadrature recipe~\cite{petras1999computation}, a cable arrangement plan stage was produced which could provide an advanced 3D cable model. Under similar given limits with simulation, a real cable design was measured through {LS}.
\section{Mobile Laser Scanning ({MLS})}
When it comes to using {MLS} point clouds for creating 3D models, the current methods usually demonstrate great accuracies, but are not yet fully autonomous, and require high computational costs, i.e. there exist no ideal methods as all of the methods entail their own drawbacks, which will be discussed in what follows.
\subsection{Extracting Road Information}
{MLS} point clouds can be utilized to extract road information, whose examples will be provided in what follows.
\subsubsection{3D Local Feature {BKD}}
Autonomous vehicle navigation, which requires obtaining road information from point clouds, is one of the applications of {MLS}, which can be achieved through {BKD}s, consisting of Gaussian kernel estimation and binarization~\cite{yang20173d}. More clearly, the curbs and markings on the road are detected through {RF} classifiers, from the shape and intensity data present in the point clouds, being then processed in order to extract the number of lanes, as well as their widths and intersections. However, point density and noise are among issues that might affect the performance of such a system, which have been tackled with an accuracy of around 90\% in the aforementioned study.
\subsubsection{Semi-automated Delineation}
In~\cite{yang2013semi}, a method was proposed for the extraction and delineation of 3D roads utilizing the {GPS} time, in order to separate the point clouds of the {MLS} system into a set of sequential road cross-sections and then filtering the non-ground points by applying a window-based filtering operator. Three types of curbs, namely, jump, point density, and slope change, were modeled and integrated into the proposed method, which detects the curb points from filtered point clouds. Visual inspection and quantitative evaluation showed that the proposed method is effective at extracting 3D roads from {MLS} point clouds, even in complex urban street-scenes. The disadvantage of the foregoing method is that it is difficult to deal with curbs with boundaries that are characterized as asphalt/soil, asphalt/vegetation, or asphalt/grassy bank.
\subsubsection{Detecting Road Boundaries}
In~\cite{zai20163d}, an algorithm was proposed for extracting road boundaries from {MLS} point clouds. The algorithm was tested on a point cloud database acquired by a RIEGL VMX-450 system~\cite{riegl450}. In general, the algorithm performed non-ground point removal by creating a set of grids based on separating the space into voxels in the vertical direction.  Afterward, the lowest points with highest density are selected. In the next step, the road curbs are detected using energy minimization and graph cuts, which may fail on roads with waved surfaces or roads without curbs.
\subsection{Street Object Recognition}
On top of road information, data about other objects present in the scene can be extracted using {MLS}, which will be discussed in what follows.
\subsubsection{Feature Matching}
In~\cite{4sigvox}, an approach was proposed for urban road object recognition from {MLS} point clouds, namely, robust 3D multi-scale shape descriptors referred to as {SigVox}, which is based on eigenvectors' properties and recursive subdivision of each potential point cluster using the octree algorithm~\cite{meagher1982geometric}. Significant eigenvectors of the points in each voxel are determined by {PCA}, and mapped onto the appropriate triangle of a sphere approximating an icosahedron. The latter step is repeated for different scales.\\
The number of required levels of the {SigVox} descriptor depends on the complexity of the geometric shape of the selected objects of interest. The proposed approach has been tested on 4~km road, and achieved over 94\% of accuracy compared to the ground truth data. It presents a shape descriptor for complete objects, in order to efficiently extract repetitive objects from large scene point clouds.
\subsubsection{Detecting Street Lighting Poles}
{MLS} point clouds have been utilized for the purpose of detecting street lighting poles, which consists in removing ground points from the frames returned by the RIEGL VMX-450 {MLS} system~\cite{guan2014using} using elevation-based filtering, clustering the rest of the points according to Euclidean distance, applying segmentation for separating overlapping objects by means of voxel-based {Ncut}~\cite{shi2000normalized}, and performing statistical analysis on the geometric features, so as to extract 3D representations of the lighting poles~\cite{zai2015inventory}.
\section{Terrestrial Laser Scanning ({TLS})}
{TLS} is a major type of {LS}, whose applications will be reviewed in this section.
\subsection{Archaeological Site Documentation}
3D modeling of monuments can be achieved in a fast, accurate and flexible manner, using {TLS}. It will be discussed in what follows.
\subsubsection{Byzantine Land Walls of Istanbul}
The performance of {DSLR} cameras has been compared to classical photogrammetric dense matching of stereo images, i.e. point clouds, in the PIXEL-PHOTO software, considering a scene consisting of the Byzantine Land Walls of Istanbul as a case-study, in~\cite{kim2013digital}.
\subsubsection{The Temple of the Sacred Tooth Relic at Kandy, Sri Lanka}
In~\cite{rahrig2017sri}, the process of capturing the Temple of the Sacred Tooth has been presented using {TLS}, by means of the RIEGL VZ-400i~\cite{riegl400}.  Additionally, to aid the scanning process, the scanner was used in combination with a laptop and the software RiSCAN PRO~\cite{riscanpro} for controlling the device on site. The color information was captured using a Nikon D700 with 14~mm lens connected to the top of the scanner. The documentation of high-level details was done by hand-held structured-light scanner, namely, Artec MHT~\cite{artecmht}. The gathered scans were registered and triangulated with the scanning software Artec Studio~\cite{artecstudio}. Afterward, the surface model was denoised using the Geomagic Studio~\cite{geomagicstudio} software. The processed 3D models were used to generate drawings of buildings with AutoCAD~\cite{li2012gamicad}, and processed with the Add-on Pointsense Heritage from Faro/Kubit~\cite{fabropointsense}.
\subsubsection{Geoarchaeological Sites in Jordan, Egypt and Spain}
\begin{figure}[t]
	\centering
	\includegraphics[width=.48\textwidth]{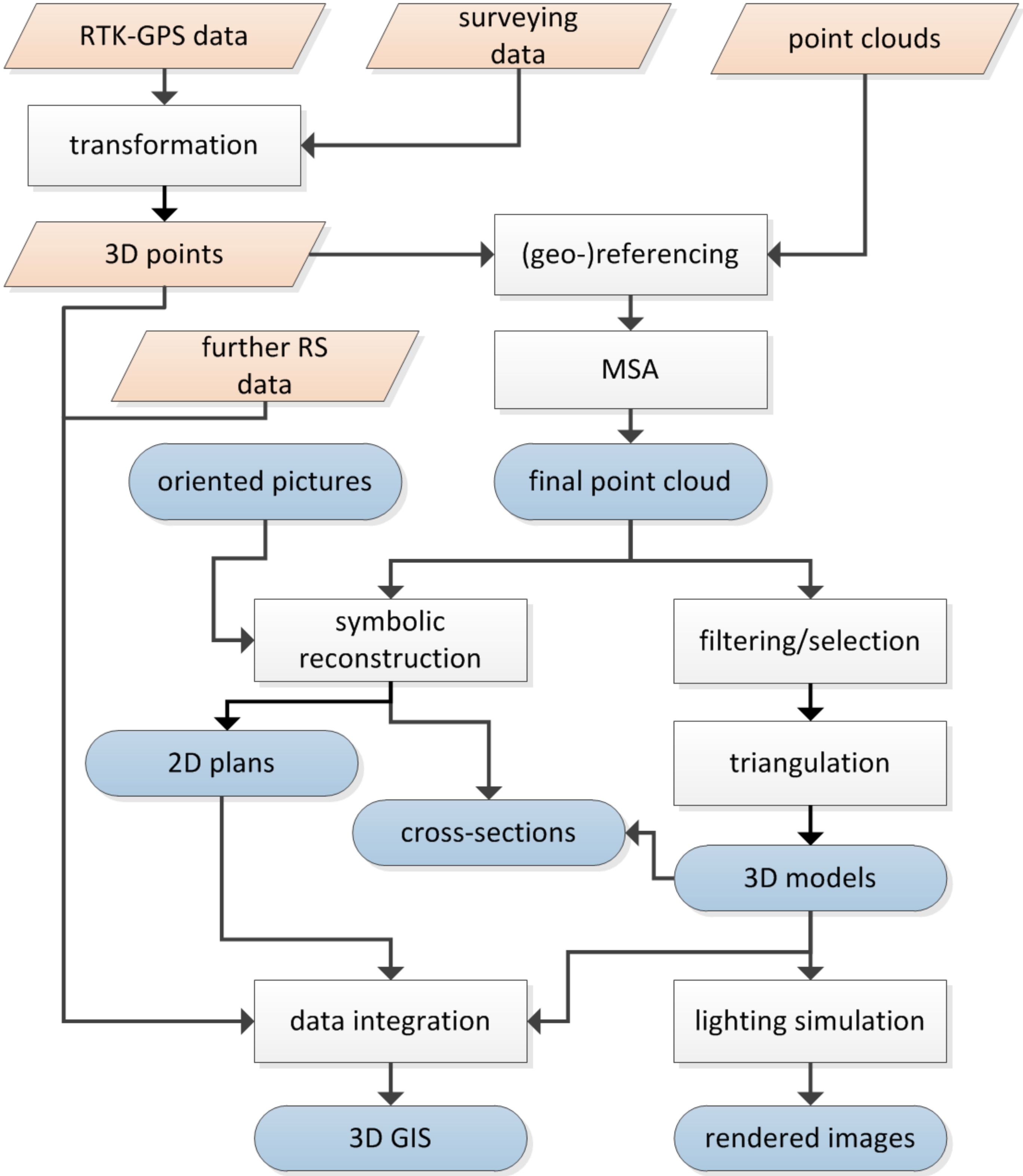}
	\caption{Schematic representation of the post-processing stages of the method proposed in~\cite{hoffmeister2014geoarchaeological}. The figure has been taken from~\cite{hoffmeister2014geoarchaeological}.}
	\label{2}
\end{figure}

Geoarchaeologically interesting sites can be documented using {TLS}, which has been done on samples in Jordan, Egypt and Spain in~\cite{hoffmeister2014geoarchaeological} using the LMS-Z420i from Riegl~\cite{schneider2009calibration} and an {RTK}-{GPS}~\cite{tamura2002measurement} for georeferencing the point clouds~\cite{harwin2012assessing}. Local surveying networks~\cite{chades2011general} were also incorporated for the sake of registration. The resulting data were then utilized for volumetric determination, approximating the ceiling thickness of caves and simulation of lighting using path tracing under harsh experimental conditions, with the weight of the device, the time-consumption and the minimum measuring distance as the underlying limitations. The results can be imported into a 3D {GIS}~\cite{koller1995virtual}. The post-processing stages from the foregoing study have been schematically illustrated in Fig.~\ref{2}.
\subsection{Data Acquisition for Indoor Assets}
In~\cite{lee20133d}, the accuracy of {TLS} was evaluated in terms of scanning a room and modeling it, through comparing the measurement differences. The {RMSE} was 2~cm. They used the Leica ScanStation C10~\cite{abbas2013calibration}, which is a high-accuracy medium-range, i.e. 0.1$\sim$300~m, laser scanner, the Leica Cyclone~7.3 point cloud processing software~\cite{leicacyclone}, which is capable of aligning multiple scans and removing the noise, and the SketchUp 3D modelling software~\cite{wolk2008utilizing}. They made the final models from point clouds manually.
\subsection{Urban Environment Modeling}
Scene modeling  is a major application of {LS}. Street, i.e. urban, modeling by means of a {HD} {LiDAR} laser scanner was proposed in~\cite{babahajiani2016comprehensive}, which enables detecting, segmenting and classifying objects present in the resulting point cloud. Binary range image processing was employed for detecting building facades, where the rest of the point cloud is segmented into voxels, and subsequently, super-voxels. After extracting features from the latter, they were classified by means of trained, boosted decision trees into semantic classes such as tree, pedestrian and car. Finally, the scene was reconstructed through rendering facades using the ShadVis algorithm~\cite{duguet2004point}, followed by fitting a prescribed solid mesh model to the aforementioned classified objects. The proposed method was qualitatively and quantitatively assessed on the {TLS} NAVTEQ database~\cite{zhou2012super}.\\
Moreover, in~\cite{yu2017application}, a viable strategy was proposed to obtain high-accuracy spatial data of urban structures. The practicability of urban structures' spatial data was discussed in view of the earthbound 3D {LS} technology. The analysis results demonstrated the legitimacy of the data. The use of the technology can diminish the field workload, avoid labor climbing, and maintain a strategic distance from touch estimation of each building, which can upgrade the proficiency.
\section{Airborne Laser Scanning ({ALS})}
According to~\cite{5comparision}, the main applications of {ALS} are in the following items:
\begin{itemize}
	\item Mapping of surfaces with very little texture definition;
	\item Mapping of forests and vegetated areas, where {ALS} can provide measurements on the ground;
	\item Mapping of long, narrow features;
	\item {DSM} generation of urban regions for urban planning and roof-top heights for communication antennas;
	\item Mapping of very small objects, e.g. power lines;
	\item Fast response applications.
\end{itemize}
Some of the above applications will be reviewed in what follows.
\subsection{Landslide Activity Analysis}
In~\cite{11rock}, geospatial methods were utilized in order to analyze landslice surface of complex rockslide systems, which are affected by strong surface pattern changes. 3D vectors and displacement fields were detected using {ALS}, based on image correlation and breakline tracking. Image correlation algorithms used in the foregoing study are fully automated and less time-consuming than breakline tracking. However, image correlation usually produces more accurate results. In either case, breakline tracking is currently the only method which allows to detect displacement vectors for areas with very high surface pattern changes.
\subsection{Forest Canopy Analysis}
Forest canopy data have been broadly obtained using {ALS} in the literature, which will be illustrated in what follows.
\subsubsection{Quantification}
{ALS} can be utilized for reconstruction of 3D forest canopy, which has been done with a density of 0.6-0.8 points.m$^{-2}$ and a resolution of 400-900 pixels.m$^{-2}$ in~\cite{vauhkonen2015reconstruction}. It can be performed based on topological connectivity, computational geometry and numerical optimization~\cite{da20133d}, with ordered sets of simplices as outputs. In the context of {ALS}, triangulation consists in subdividing the point space into weighted simplicial complexes, hence quantifying it, requiring filtering out the canopy voids. In~\cite{vauhkonen2015reconstruction}, vegetation point clouds were obtained while optimizing the filtration based on field measurement and analyzing the persistent homology, thereby calculating the tetrahedral volume with relatively high degrees of determination~(R2), i.e. 0.65 alone and 0.78 together with the rest of the predictors. The field measurements were reported to enhance the homology persistence of the point clouds representing the stem volumes.
\subsubsection{Estimating the {LAI}}
The {LAI} is one of the main metrics used when quantifying the energy exchange between land surfaces and the atmosphere. In~\cite{boreal_air_2}, an approach was proposed to estimate forest canopy effective {LAI} by 3D triangulation of {ALS} point clouds. The benefit of their approach over the more commonly used Beer's law-based approach~\cite{thorne1993comparison} is that the triangulation also produces shape estimates.
\section{{TOF}, Structured Light and Phase Comparison}
Due to their underlying comparatively simple technology, {TOF} sensors bear potential for economical and handy scanning procedures, which will be illustrated in this section. It should be noted that turntable-based 3D scanning requires a specifically designed tool or markers for calibration, since otherwise, the rotation axis will be lost after the space configuration is changed. Moreover, the accuracy of the inter-beam points in the anticipated optical pattern impacts the precision of the framework.
\subsection{Motion Compensation}
In~\cite{10motion}, a 3D scanning system combining stereo and active illumination based on phase-shift was proposed for robust and accurate scene reconstruction. The system works in real-time, and aims at constructing complex scenes of several independently moving objects at 17 {FpS}. In stereo phase-shift, correlation-based calculations are utilized with optimization at segment level. It allows to avoid the problem of discontinuity. To overcome the motion of artifacts at pixel level, they applied an analysis of the motion error.
\subsection{A Tool-free Calibration Method}
In~\cite{pang2016tool} a gray-coded structured-light pattern projected onto the object was used for calibration, which outperforms a sphere-based calibration model both in terms of accuracy and ease of use. However, it should be noted that the foregoing method fails with reflective, featureless and slippable untextured objects.
\subsection{Improved Coverage through Online Structured-light Calibration}
Structured-light 3D scanners suffer from incomplete coverage problems due to limited lines of sight, which has been alleviated in~\cite{hc3d}, through a novel online method by augmenting the use of a pair of stereo cameras coupled with a non-calibrated projector. The main advantage of this method is that it can be performed during the normal scanning process, allowing an improved scene coverage with little or no additional effort.
\subsection{3D Shape Scanning with {TOF} Sensors}
In~\cite{cui2013algorithms}, it was demonstrated that 3D shape models of static objects can be acquired with a {TOF} sensor, which was achieved by the effective combination of 3D super-resolution with a new probabilistic multi-scan alignment algorithm tailored to {TOF} sensors, by aligning depth scans that were taken from around an object with a {TOF} sensors at video rate. The algorithmic challenge they faced was that the sensor's level of random noise was substantial, along with a nontrivial systematic bias.
\subsection{Low-cost Hand-held 3D Scanning of Architectural Elements}
In~\cite{allegra2016low}, the feasibility of using hand-held structure sensor scanners for creating 3D meshes of architectural objects was investigated. A structure sensor is a near-{IR} structured-light 3D device that works well for scanning objects of volumes less than 1~m$^3$. To test the scanner's ability to scan larger objects, an 18$^\text{th}$ century doorway in Benedettini monumental complex in Catania~(Sicily) was scanned with a structure sensor and {TOF} laser scanner, namely, HDS 3000 by Leica Geosystem~\cite{reshetyuk2006calibration}, which was used as a reference model.  Due to the limited precision of the structure sensor, the doorway was scanned in 23 segments, with 30\% overlap, and afterward, put together. The results showed that the mean distance from the {TOF} model for details was less than 5~mm and 9.6~mm for the whole model, thereby demonstrating that hand-held structure sensors are affordable and precise solutions for digitizing cultural heritage.
\subsection{The Kinect Sensor}
The Kinect sensor~\cite{Kinect} has greatly reduced the underlying computational and financial costs of making 3D models of objects and persons, which will be discussed in what follows.
\subsubsection{Objects~\cite{dou20153d}}
In~\cite{dou20153d}, an algorithm was developed for making 3D scans of deformable objects and people without requiring access to prior information or having them stay still. The method uses a single Kinect sensor~\cite{Kinect}. They detect loop-closures to handle severe drifts, and perform dense nonrigid {BA}, in order to correct the evenly distributed error from a pure loop-closure. The final result is a unified 3D model that does not need an initial template. The method is susceptible to complex changing topologies in the scene, and the computational cost is relatively high: With a 8-core 3.0G Hz Intel Xeon {CPU} and 64~{GB} of memory, a sequence with 400 frames takes 9$\sim$10~hours.
\subsubsection{Persons}
3D scanning and printing procedures can be utilized for creating miniatures of persons, which has been handled by means of a Kinect sensor~\cite{Kinect} and a color printer in~\cite{sturm2013copyme3d}, where color and depth frames are acquired while the person is rotated on a swivel chair. The model is created, updated and visualized in real-time through adopting a {SDF}~\cite{bylow2013real}, and possible holes caused by self-occlusions are automatically filled. In order to minimize the 3D printing costs, a watertight but hollow shell is then obtained, where the quality of the resulting miniature depends on numerous factors, including the camera pose and the consequent self-occlusions, the non-rigid deformations of the subject's body and the rotation speed.
\subsection{Polarization and Phase shifting}
Polarization and phase-shifting approaches have been utilized to a considerable extent for the sake of developing 3D scanners, whose examples will be provided in this section.
\subsubsection{Translucent Objects}
Performing 3D scanning on translucent objects is considered a challenging task, which is prone to errors, as subsurface scattering shifts the intensity peak beneath the surface of the object. In~\cite{chen2007polarization} phase shifting and polarization filtering were tested, and then combined. However, it should be noted that they only tested the algorithm on homogeneous, i.e. single-color, single-material, objects.
\subsubsection{A Wideband Antenna}
In~\cite{9antenna}, a design of a wideband \textit{45} polarized electrical scanning antenna array was proposed for metallic 3D printing. As wideband array elements, they used a ridged horn. To avoid the problem of increase in the element space in the scanner plane, they designed a rigged horn of `S' shape and a
bulkhead between the two ridges. The last modification makes array suitable for $\pm$\textit{30} scanning. The test of the proposed antenna array resulted in a less than 0.3~dB measured loss in 3D printing.
\subsection{An Infrared System for Metallic Surfaces}
In~\cite{6metallicsurfaces}, an application of {SfH} was proposed, namely, an operational 3D digitizer for metallic objects. They have shown that laser and an {IR} camera can be set up to digitize metallic specular surfaces in 3D. Compared to methods based on polarization imaging or deflectometry, the {SfH} method proposed in the aforementioned study results in higher measurement errors, but is applicable to much larger sets of surfaces. Also compared to systems developed in the visible domain, {SfH} is less dependent on the roughness of the surface.
\section{Comparative Studies}
Due to the fact that 3D scanning methods and apparatuses constitute a significantly wide range, numerous studies have attempted to compare them from various perspectives, a few examples of which will be reviewed in what follows.
\subsection{3D Scanning Instruments}
In~\cite{boehler20023d}, a general overview of {TOF} and phase comparison 3D scanning technologies has been provided. As there are no instruments that can be applicable to all scanning situations, the scanning devices are divided into three general range categories: 0.1 to 1, 1 to 10 and 10 to 100 m. When considering the practical use of 3D scanning equipment, the speed of point acquisition can increase or decrease the working time for scanning an area or object. Typically, higher scanning resolutions are preferred. Each scanner is limited by its {FoV}, with typical scan area of about \textit{40} by \textit{40}. Large {FoV}s have the potential to collect large amounts of data from a single observation. In addition to the distance measurements, users may desire photo-realistic views of scanned objects, which can be achieved by adding high-quality cameras to the scanning setup. Ease of transportation is a major factor, i.e. scanners preferably should be small and lightweight, and have portable generators, in addition to power line supply. Last but not least, the scanning software must be fast and simple. It should make dynamic adjustments, and calculate the scanning times.
\subsection{A Comparison of Photogrammetry and {ALS}}
In~\cite{5comparision}, data acquisition and processing from passive optical sensors, i.e. photogrammetry, and {ALS} were compared. The main differences between them are overviewed in Table~\ref{xxx}.
\begin{table*}[ht]
	\caption{The main differences between photogrammetry and {ALS}.}
 	\resizebox{\textwidth}{!}{
		\begin{tabular}{|c|c|}
			\hline 
			Photogrammetry                                                                                                                                                                                                                                     & {ALS}                                          \\ \hline\hline
			Passive                                                                                                                                                                                                                                                     & Active                                                \\ \hline
			Frame or linear sensors with perspective geometry                                                                                                                                                                                                            & Point sensors with polar geometry                     \\ \hline
			Full area coverage                                                                                                                                                                                                                                          & Pointwise sampling                                    \\ \hline
			Indirect acquisition                                                                                                                                                                                                                                        & Direct acquisition                                    \\ \hline
			Geometrically and radiometrically high-quality images with multispectral capabilities                                                                                                                                                                       & No imaging or monochromatic images of inferior quality \\ \hline
			Height of flight over 1000                                                                                                                                                                                                                                & Height of flight over 6000                          \\ \hline
			
			\begin{tabular}[c]{@{}c@{}c@{}}High variability: 2D, linear and multiple-line methods and point detectors\\Geometric, radiometric and spectral resolution\\Geometric accuracy measures, i.e. metric, semi-metric and non-metric\end{tabular}
			 & Less variability and flexibility                      \\ \hline
			Higher flying speed and height                                                                                                                                                                                                                               & Lower flying speed and height                          \\ \hline
			Requiring manual intervention                                                                                                                                                                                                                                   & Under ideal conditions, proved fully automatic \\ 
			\hline
		\end{tabular}
 	}
	\label{xxx}
\end{table*}
\subsection{Fast Methods for {LS}}
3D perception has various applications in manufacturing, automatic control and robotics. A review of {TOF} 3D measurement and scanning systems has been provided in~\cite{wulf2003fast}, evaluating the performance of the relevant techniques based on the number of points captured from an area of interest within a given time interval, which has resulted in recommendations as to how to improve the efficiency of mechanical scanning devices in terms of speed and accuracy. In the aforementioned study, the methods taken into account have been compared through implementing them on a mobile robot with a Linux operating system, being run in real-time.
\subsection{Building and Urban-area Modeling}
As demonstrated in the previous sections, buildings and urban areas can be scanned in 3D using a variety of methodologies. Some of the studies comparing the related approaches in terms of their performance in the foregoing context will be discussed in this section.
\subsubsection{Heritage Sites}
In~\cite{remondino2011heritage}, a review of a diverse range of optical 3D measurement and modeling devices, techniques and algorithms developed by the time has been provided, which can be considered as a reference when it comes to focusing specifically on the progress made during the last few years by means of comparing the state-of-the-art on the time the aforementioned article has been written with the one as of now, in terms of the underlying properties, limitations, capabilities and limitations.
\subsubsection{A Comparison of Aerial Photogrammetry and {LiDAR}}
{DSM}s resulted from stereoscopic aerial images constitute a means of change detection or 3D reconstruction from urban areas. In~\cite{stal2013airborne}, they have been qualitatively and quantitatively compared with airborne {LiDAR} ones, where three areas in the city center of Ghent, Belgium, have been considered as case-studies, each of which covers around 0.4 km$^{2}$. The analysis relied on pixel-wise comparison of the models, where the real changes were distinguished from false alarms caused by noise, vegetation or outliers. The newly reconstructed or removed buildings were detected based on the difference models, whose surfaces and volumes were quantified as well.
\subsubsection{Terrestrial Photogrammetry for Complex Buildings}
In~\cite{2eval}, different software for photogrammetric 3D modeling were compared. The models were compared with {LS}. The software considered were Agisoft Photoscan~\cite{verhoeven2011taking}, MicMac~\cite{micmac}, Bentley ContextCapture~\cite{contextcapture} and Visual{SFM}~\cite{wu2011visualsfm}, which were evaluated in terms of modeling buildings. For comparison, the {C2C} distance was considered as the criterion. From selected buildings, 150 photos were taken. Photoscan and ContextCapture returned results of comparable accuracy with that of {TLS}, but overall, Photoscan led to a better understanding of the geometry. MicMac demonstrated an accurate representation of the geometry as well, but appeared noisy, with some gaps. Visual{SFM} produced the noisiest results, with a weak geometric performance.
\subsection{Unmanned Aerial System ({UAS})}
A review of the applications of {UAS}s in {PaRS} has been provided in~\cite{colomina2014unmanned}. Including among others, the technical aspects, e.g. sensing, navigation, steering or data processing, privacy-related issues, safety, and war and peace topics as regards {RPAS} such as {UAV}s, aircrafts and drones have been investigated, along with the consequent conflicts with the traditional aeronautical methods and regulations, paying a particular attention to the ones operating at nano, micro or mini scales.
\subsection{Forest or Vegetated-area Analysis}
Forests and vegetated areas can be 3D-scanned using various approaches, which have been compared in the studies reviewed in what follows.
\subsubsection{A Comparison of Low-altitude {UAV} Photogrammetry with {TLS} for Terrain Covered in Low Vegetation}
The results of performing photogrammetry using {UAV} and {TLS} have been combined in~\cite{gruszczynski2017comparison}, in order to explore and detect land relief, requiring rather high precisions for mapping of the land's natural and anthropogenic uplifts and subsidences. The possible effect of vegetation on the detection of the surface has been reported as one of the underlying challenges, tackling which demands determining and excluding the areas whose heights are severely affected by the foregoing issue. In~\cite{gruszczynski2017comparison}, the latter task was handled through capturing dense measurements returned by a tacheometer and a rod-mounted reflector. As an affirmative conclusion, it has been realized that {UAV} is more efficient than {TLS} for the purpose of photogrammetry.
\subsubsection{Predicting the {SI} in Boreal Forests Using {ALS} and Hyper-spectral Data}
In sustainable forest management, the {SI} is the most common quantitative measure of site productivity, which is usually determined for individual tree species based on tree height and the age of the largest trees. In~\cite{boreal_air}, a method has been proposed to determine the {SI} using {RS} data. In particular, they fused on {ALS} and airborne hyper-spectral data, based on {ITC} delineation, i.e. the tree species and height, {DBH} and age were modeled and predicted at {ITC} level. These four dominant {ITC}s per 400~m$^2$ plot were selected as inputs in order to predict the {SI} at plot level.
\subsection{Digital Elevation Model ({DEM}) based Comparison}
{DEM}s can be obtained utilizing a variety of approaches. A few sample studies which have compared items from the foregoing list will be discussed in this section.
\subsubsection{A Comparison of United States Geological Survey ({USGS}), Digital Line Graph ({DLG}), Shuttle Radar Topographic Mission ({SRTM}), a Statewide Photogrammetry Program, Advanced Spaceborne Thermal Emission and Reflection ({ASTER}), Global {DEM} ({GDEM}) and Light Detection And Ranging ({LiDAR})}
The field of geomorphologic analyses relies on topographic changes calculated from {DEM} differencing. Over the years, many such models were proposed, which differ greatly in terms of the collection procedure, resolution, and accuracy level. The study reported in~\cite{dems_comp} compares several regional- and global-scale {DEM}s with a high-accuracy {LiDAR} {DEM}, in order to quantitatively and qualitatively assess their differences in the rugged topography of the southern West Virginia, USA coalfields. They concluded that {SRTM} {DEM}s and {GDEM}s describe a topographic surface substantially preferable to the one modeled by {LiDAR} {DEM}.
\subsubsection{Quality Assessment of {TanDEM-X} Using Airborne {LiDAR} and Photogrammetry}
In~\cite{tandem_dems}, a survey was provided of {DEM}s gathered using {TanDEM-X}, which is {TSX}'s twin, a German Earth observation satellite using {SAR} technology, where interferometric data are acquired in bistatic mode. Specifically, {TanDEM-X} was evaluated in comparison with other mainstream {DEM}s such as {LiDAR}. The study concluded that {TanDEM-X} raw {DEM}s are comparable, with {RMSE}s being around 5~m, to {LiDAR} ones.
\subsubsection{A Comparison of {LiDAR} and Photogrammetry in Tenerife Island}
{DTM}s representing ground elevations can be obtained through manual photogrammetric restitution of altimetric features~\cite{toutin1995real} or aerial {LiDAR}~\cite{verma20063d}, which have been compared on the basis of various vegetation covers in the island of Tenerife, Canary Islands, Spain, in~\cite{gil2013comparison}, both with pixel sizes of 5~m, and the latter possessing a nadir density of 0.8 points.m$^{-2}$. The foregoing study analyzed three elevation profiles through nonparametric methods, utilizing a {TS}, which led to an accuracy of between 0.54~m and 24.26~m for the photogrammetry-based approach and between 0.22~m and 3.20~m for the {LiDAR}-based one, at the 95\textsuperscript{th} percentile, i.e. the latter has resulted in more precise representations of the actual surfaces, especially in the case of the parts which have been invisible to photogrammetry, e.g. Canarian pine forests.
\section{Combinational Studies}
For the sake of alleviating the problems from which the 3D scanning techniques presented in the previous sections suffer, some studies have suggested devising combinational approaches. A few examples from the foregoing list will be reviewed in what follows.
\subsection{Forest Inventory Attribute Estimation Using {ALS}, Aerial Stereo Imagery, Radargrammetry and Interferometry}
The first results of using digital stereo imagery and spaceborne {SAR} in forest inventory attribute mapping in Finland have been reported in~\cite{holopainen2015forest}, on areas where detailed {ALS}-based {DTM} is available. Alternative 3D techniques such as aerial imagery and Spaceborne {SAR} have been utilized to study inventory attribute estimation in three different study sites in Finland. The results were also compared to {LiDAR}-based {ALS}.\\ 
From the user's point of view, the preference on the aforementioned 3D {RS} methods in forest inventory attribute updates depends on data availability, acquisition and preprocessing costs. User-ready products such as {DSM}s or point clouds can be easily processed with available software, and used for area-based estimation of forest inventory attributes.
\subsection{Cultural Heritage Documentation}
A few combinational methods which have been proposed for the purpose of documenting cultural heritage will be discussed in what follows.
\subsubsection{Metric Documentation of Cham Towers in Vietnam Using {LS} and {PSP}}
3D reconstructions of Cham towers in Vietnam have been presented in~\cite{fangi2013metric}. The models were made using {LS}, {SFM} and {PSP}. {LS} was performed using Z+F IMAGER 5006h~\cite{mechelke2007comparative}, and processed by means of the Cyclone software~\cite{frohlich2004terrestrial}. The {SFM} algorithm was applied using the Agisoft Photoscan software~\cite{verhoeven2011taking}. Finally, {PSP} models were created using two Cannon cameras with resolutions of 12 and 14~Mega-pixels. The obtained images were stitched together with the PTGui~9~\cite{ptgui} software, and the model formation was created using the Sphere package and bundle block adjustment~\cite{kerschner1998homologous}. The authors preferred the {PSP} procedure because of the ease of use, although the {LS} system resulted in more precise model reconstructions.
\subsubsection{A Comparison of Digital Photogrammetry and {LS}}
In~\cite{yastikli2007documentation}, technologies related to digital photogrammetry and {TLS} were discussed, along with monoscopic multi-image evaluation, stereo digital photometry and {TOF} {TLS}. Monoscopic multi-image evaluation methods were used to document historical building facades. The task of image acquisition was performed using a Nikon D100 {DSLR} camera~\cite{nikond100}. The creation of 3D photogrammetric line drawings was carried using efficient {CAD} functionalities provided by the MicroStation software~\cite{zlatanova2004topological}. Stereo digital photogrammetry was used for documenting the Fatih Mosque facades, and model generation was achieved with the Z/I Phodis ST30 software. Finally, a {TLS}-based survey of the Muayede~(Ceremonial Hall) of the Dolmabahce Palace was performed with a {TOF} scanner, namely, LMS-Z420i from RIEGL~\cite{schneider2009calibration}, with a mounted Nikon D70s calibrated digital camera~\cite{gloe2010dresden}. The {DOM}s were automatically created by the RiSCAN PRO processing software~\cite{riscanpro}.
\subsubsection{Accuracy and Block Deformation Analysis in Automatic {UAV} and {TLS}}
The conclusions made through analyzing the performance of an integrated system meant for {UAV} and {TLS} have been reported in~\cite{nocerino2013accuracy}. The archaeological site of the Roman theater in Ventimiglia, Italy, has been considered as a case-study, in order to validate the proposed evaluation framework. The drawings were created at a scale of 1:20. A {GSD} of less than 4~mm was considered in order to abide by the latter requirement, which means that the accuracy was 4~mm. Both vertical and oblique images were obtained by the {UAV}. On the other hand, terrestrial images were acquired aiming at capturing individual dense point clouds of vertical structures. The robustness of the photogrammetric system, i.e. the repeatability of the experiments, was examined against various network configurations, based on the presence of ground control.
\subsection{3D Change Detection at Street Levels Using {MLS} Point Clouds and Terrestrial Images}
In~\cite{qin20143d}, a method for change detection at street levels was proposed. In order to achieve the aforementioned goal, they took the following steps:
\begin{itemize}
	\item The point clouds were recorded by an {MLS} system, and processed, with the data cleaned and classified by semi-automatic means;
	\item At a later epoch, terrestrial or {MMS} images were taken and registered onto the point clouds, which were then projected on each image by a weighted window-based z-buffering method for viewing dependent 2D triangulation;
	\item Stereo pairs of the terrestrial images were rectified and reprojected between each other, in order to check the geometrical consistency between the point clouds and stereo images;
	\item An over-segmentation-based graph cut optimization procedure~\cite{li2004lazy} was carried out, taking into account the color, depth and class information, so as to compute the changed area in the image space.
\end{itemize}
The method proved invariant to light changes and robust against small co-registration errors between images and point clouds.
\subsection{Area-based Quality Control of {ALS} Models Using {TLS}}
In 3D earth modeling using {RS} techniques, one of the most significant issues is correct surface determination under dense vegetation. {ALS} {DEM} performance under high vegetation was assessed in~\cite{sefercik2015area} using {RVA} analysis. They also assessed {VHR} {ALS} 3D earth models using visual approaches and {TLS}-based reference models. The {AVA}s of {VHR} {ALS} {DSM}s in open, grass, and building areas were reported to be very similar, with 3-5~cm $\sigma$ and normalized median absolute deviation, which in high-vegetation areas were 35~cm. It also revealed that {ALS} {DEM} {AVA} under canopy satisfies the requirement for a 1/1000-scale topographic map.
\subsection{Single-shot 3D Scanning via Stereoscopic Fringe Analysis}
In~\cite{dense_scan}, a single-shot method was proposed that improves both density and accuracy of 3D scanning by combining two conventional techniques, namely, stereoscopic and Fourier fringe analyses~\cite{bone1991fourier}. In addition, due to its low complexity, the proposed method performs well even in time-critical applications.
\section{Post-processing}
Raw point clouds resulted form 3D scanning procedures usually suffer from presence of outlying points and jaggednesses in their apparent surfaces, which are caused by noise. Thus post-processing them is necessary for removing the outliers and smoothing the surface, which will be discussed in this section.
\subsection{Outlier Detection and Normal-curvature Estimation}
In~\cite{3outlier}, two techniques were proposed for outlier detection in {LS} 3D point cloud data. The first algorithm was based on {RZ-score}, which is a classical distance-based measure computed as follows:
\begin{align}
	Rz_i = \frac{p_i - \text{median}(p_i)}{MAD},
\end{align}
where $MAD$ stands for {MAD}. If the calculated Rz-score for observation $z_i$ is greater than or equal to 2.5, then it is deemed an outlier.\\
The second algorithm uses {MD} for outlier detection, which is defined as follows:
\begin{align}
	MD=\sqrt{(p_i-\bar{p})^T\sum^{-1} (p_i-\bar{p})}
\end{align}
where $\bar{p}$ and $\sum{}$ are sample mean and covariance matrix, respectively. An observation that {MD} score exceeds 3.075 will indicate an outlier. Both of the algorithms first fit the best plane
based on the majority of consistent data or inliers within the local
neighborhood of each point of interest. Then based on the majority of the acceptable points, the outliers are defined locally for each neighborhood.
\subsection{Robust Statistical Approaches}
\begin{figure}[t]
	\centering
	\includegraphics[width=.48\textwidth]{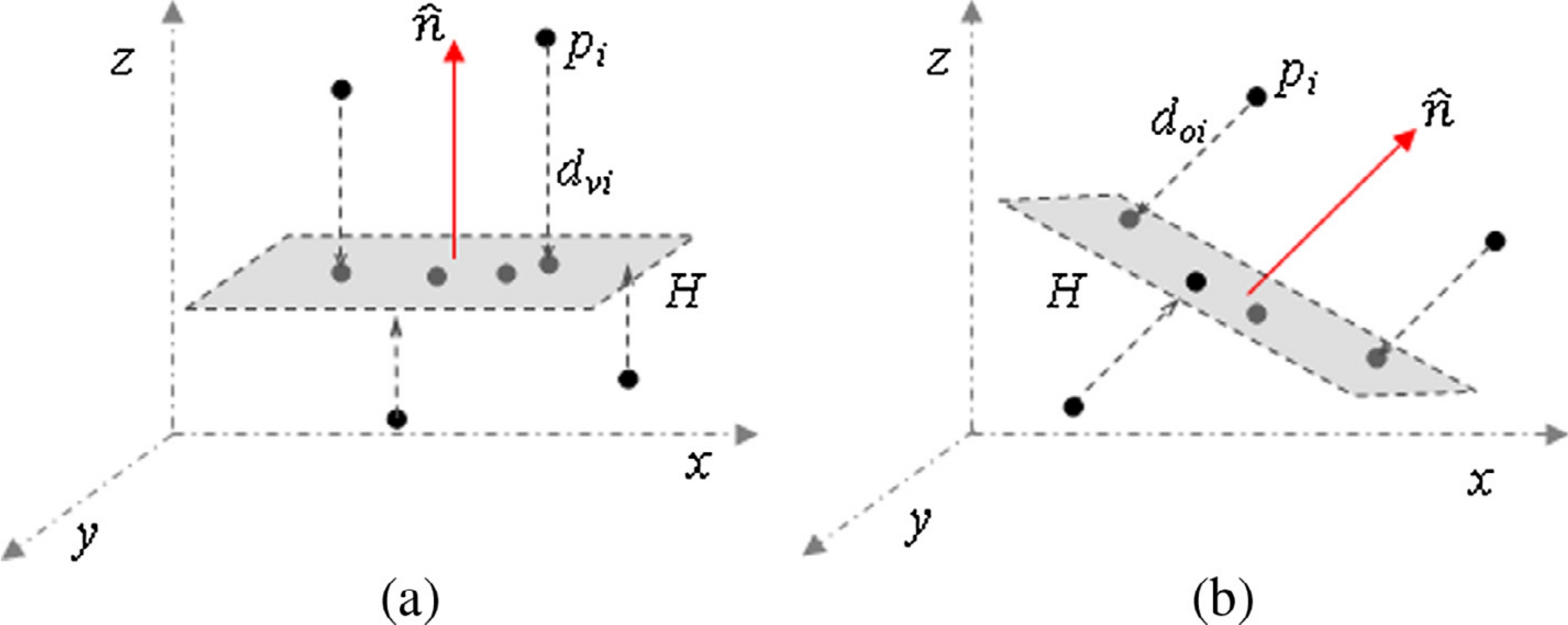}
	\caption{Comparison of (a)~least squares and (b)~total least squares, for fitting planes and estimating normals. The figure has been taken from~\cite{nurunnabi2014robust}.}
	\label{3}
\end{figure}

The focus of~\cite{nurunnabi2012diagnostic} is planar surface fitting and local normal estimation of a fitted plane. The proposed Diagnostic-Robust Principal Component Analysis ({DRPCA}) algorithm is a combination of diagnostics and robust statistical techniques. Initially, candidate outliers are found using {RD}, which reduces some outlier effects, and makes the data more homogeneous. Afterward, Robust PCA ({RPCA}) is used to find more candidate outliers \cite{croux2007algorithms}, and fit the plane. As far as plane fitting for point clouds is concerned, the proposed method outperforms least squares, {PCA}, Demixed-PCA, {MSAC} and {RANSAC} on simulated and real databases.\\
In~\cite{nurunnabi2014robust}, robust methods were proposed for local planar surface fitting of 3D {LS} data by focusing on the Deterministic Minimum Covariance Determinant estimator~\cite{hubert2010minimum} and {RPCA}, and by using variants of statistically robust algorithms. Fig.~\ref{3} illustratively compares least squares and total least squares in fitting planes and estimating normals.
\section{Conclusion}
This paper reviewed the existing 3D scanning methods, including close-range, aerial, structure-from-motion and terrestrial photogrammetry, and mobile, terrestrial and airborne laser scanning. time-of-flight, structured-light and phase-comparison approaches were covered as well, followed by comparative and combinational investigations and outlier detection and surface fitting strategies. The conclusions derived from this paper will have substantial implications regarding deciding on the best possible plan for performing the task of 3D scanning.
\section*{Acknowledgement}
This work has been partially supported by Estonian Research Council Grants (PUT638), The Scientific and Technological Research Council of Turkey (T\"UBiTAK) (Proje 1001 - 116E097) , the Estonian Centre of Excellence in IT (EXCITE) funded by the European Regional Development Fund and the European Network on Integrating Vision and Language (iV\&L Net) ICT COST Action IC1307.

\bibliographystyle{IEEEtran}
\bibliography{3DScanning}

\end{document}